\documentclass[11pt,a4paper]{article}
\usepackage{lmodern}
\usepackage[utf8]{inputenc}
\usepackage[T1]{fontenc}
\newcommand{\emoji}[1]{\includegraphics[height=0.75em]{#1}}
\usepackage{latexsym,times}
\usepackage{url}
\usepackage{textcomp}
\usepackage{wasysym}
\usepackage{graphicx}
\usepackage[table]{xcolor}
\usepackage{booktabs}
\usepackage{placeins}
\usepackage{todonotes}
\usepackage{microtype}

\usepackage[acceptedWithA]{tacl2021v1}

\usepackage{xspace,mfirstuc,tabulary}

\newif\iftaclinstructions
\taclinstructionsfalse 
\iftaclinstructions
\renewcommand{\confidential}{}
\renewcommand{\anonsubtext}{(No author info supplied here, for consistency with
TACL-submission anonymization requirements)}
\newcommand{\instr}
\fi

\iftaclpubformat 

\else

\fi

\usepackage{color}

\usepackage{dblfloatfix}

\title{Tik-to-Tok: Translating Language Models One Token at a Time: \\ An Embedding Initialization Strategy for Efficient Language Adaptation}

\author{François Remy\Thanks{François and Pieter are shared first authors}$~~^{\diamond}$, Pieter Delobelle$^{*~\dagger}$, Bettina Berendt$^\dagger$, Kris Demuynck$^{~\diamond}$, Thomas Demeester$^{~\diamond}$\\
\texttt{francois.remy@ugent.be}~~~~\texttt{pieter.delobelle@kuleuven.be} \\ 
$^\diamond$ IDLab (Internet and Data Science Lab), Ghent University - imec \\
$^\dagger$ Department of Computer Science, KU Leuven; Leuven.AI \\
}

\date{}

\begin{document}
\maketitle
\begin{abstract}
Training monolingual language models for low and mid-resource languages is made challenging by limited and often inadequate pretraining data. In this study, we propose a novel model conversion strategy to address this issue, adapting high-resources monolingual language models to a new target language. By generalizing over a word translation dictionary encompassing both the source and target languages, we map tokens from the target tokenizer to semantically similar tokens from the source language tokenizer. This one-to-many token mapping improves tremendously the initialization of the embedding table for the target language. We conduct experiments to convert high-resource models to mid- and low-resource languages, namely Dutch and Frisian. These converted models achieve a new state-of-the-art performance on these languages across all sorts of downstream tasks. By reducing significantly the amount of data and time required for training state-of-the-art models, our novel model conversion strategy has the potential to benefit many languages worldwide.
\end{abstract}

\section{Introduction}

Large pre-trained language models have emerged as a 
standard approach in NLP \citep{devlin2019bert,liu2019roberta,brown2020gpt}. 
Unfortunately, high-quality monolingual models exist only for a handful of languages. Worse, they are rarely kept up-to-date because of the cost of retraining.


Multilingual models, often touted as a solution, 
have their own challenges: 
interference between languages, poor tokenization, large model size, and more \citep{wang-etal-2020-negative,pires-etal-2019-multilingual}. Moreover, translating inputs into English and using a monolingual English model for inference can sometimes surpass the performance of a fine-tuned multilingual model \cite{artetxe-etal-2023-revisiting}. 

A more promising strategy seems to be model conversion. In this approach, the original tokenizer of an existing monolingual model is discarded in favor of an entirely new vocabulary, adapted to the target language; tokens shared between the two tokenizers keep their existing embedding, while newly-introduced tokens are randomly initialized \cite{artetxe-etal-2020-translation,de-vries-etal-2021-adapting,garcia-etal-2021-towards,gogoulou-etal-2022-crosslingual}.

While tokenizer upgrades for language models are nothing new, existing approaches only consider tokens as black boxes, ignoring the character strings they represent, as well as their semantics. They are thus ill-equipped to deal with languages which form multiword compounds, such as Dutch or German. In these languages, compound words can be formed by agglutinating several existing words together
(e.g. \textit{corporate credit} translates to 
the compound 
\textit{bedrijfskrediet} in Dutch).

Compounds complicate the use of the popular merge-based tokenizers, 
as letters at subword boundaries often get 
merged to the incorrect segment, resulting in several partial tokens with extra or missing letters, all representing the same concept but in different compounds (see Table \ref{fig:Tokens}). 
This reduces the amount of training data which each of these tokens receives during training, lowering the quality of their representation, thereby occupying precious space in the embedding table for low-quality tokens. This also makes them difficult to align with existing source language tokens using the trivial mapping strategies described thus-far.

%


In this work, we address this issue 
by initializing the embedding of each token of a new target vocabulary using a weighted combination of embeddings of similar tokens from the source model, the discovery of which consists of a novel ``token translation'' task which relies on the character composition and semantics of these tokens, which we approximate using the character n-grams these tokens contain.

\newpage

\newcommand{\newhline}{\newline\hspace*{0.25cm}}
\newcommand{\tokensum}[4]{{$E_{t}$[\textit{#1}] = \newline
             {\footnotesize\hspace*{0.25cm}+ 0.5 * $E_{s}$[\textit{#2}] \newhline+ 0.3 * $E_{s}$[\textit{#3}] \newhline+ 0.2 * $E_{s}$[\textit{#4}]}}}

\begin{table}[t]
    \centering
    \begin{tabular}{p{0.44\linewidth}|p{0.44\linewidth}}
        \toprule
        \multicolumn{2}{c}{\cellcolor{black!2}\textbf{\textsc{Tokens with extra letters}}} \\
        \midrule
        \tokensum{...\underline{ings}bedrijf}{...company}{...Company}{Company} & \tokensum{universiteit\underline{s}...}{university}{University}{...University} \\
        \midrule
        \multicolumn{2}{c}{\cellcolor{black!2}\textbf{\textsc{Tokens with missing letters}}} \\
        \midrule
        \tokensum{...oeding}{...feeding}{...eding}{...breeding} & \tokensum{inschrijf...}{inscribed}{inserting}{inline} \\
        \midrule
        \multicolumn{2}{c}{\cellcolor{black!2}\textbf{\textsc{Tokens with novel splits}}} \\
        \midrule
        \tokensum{Administr...}{Administ...}{...Administ}{administr...} & \tokensum{Afrik...}{Afric...}{Africa}{African} \\
        \midrule
        \multicolumn{2}{c}{\cellcolor{black!2}\textbf{\textsc{Tokens for word endings}}} \\
        \midrule
        \tokensum{...ventie}{...vention}{...inence}{invention} & \tokensum{...geerd}{...inated}{...urized}{...itized} \\
        \bottomrule
    \end{tabular}
    \caption{Categories of Dutch tokens not covered by a translation dictionary, and their corresponding English tokens based on the Tik-to-Tok strategy.\vspace*{-0.15cm}}
    \label{fig:Tokens}
\end{table}

We hypothesize and show that this strategy yields considerable advantages over training a language model from scratch, with regards to both training efficiency and downstream performance. 
In the next chapters, we develop the following two key contributions:
\begin{itemize}
    \item We propose a novel cross-lingual model conversion strategy for low-resource languages that does not require further pre-training on a downstream corpus, by leveraging a translation dictionary instead of a corpus (\autoref{ss:contrib-1}).
    \item We show that, even for mid-resource languages for which enough data exists for training dedicated language models, applying our strategy and finetuning on the available corpus performs better than training such a language model from scratch 
    (\autoref{ss:contrib-2}).
\end{itemize}

In the next section, we introduce the state of the art strategies for cross-lingual transfer learning and tokenizer upgrades. We then follow up with a more comprehensive description of our methodology in \autoref{ss:methodology}, and showcase our results on Frisian and Dutch model conversion in \autoref{ss:results}. 

\begin{figure*}[!b]
    \centering
    \includegraphics[width=1.0\textwidth]{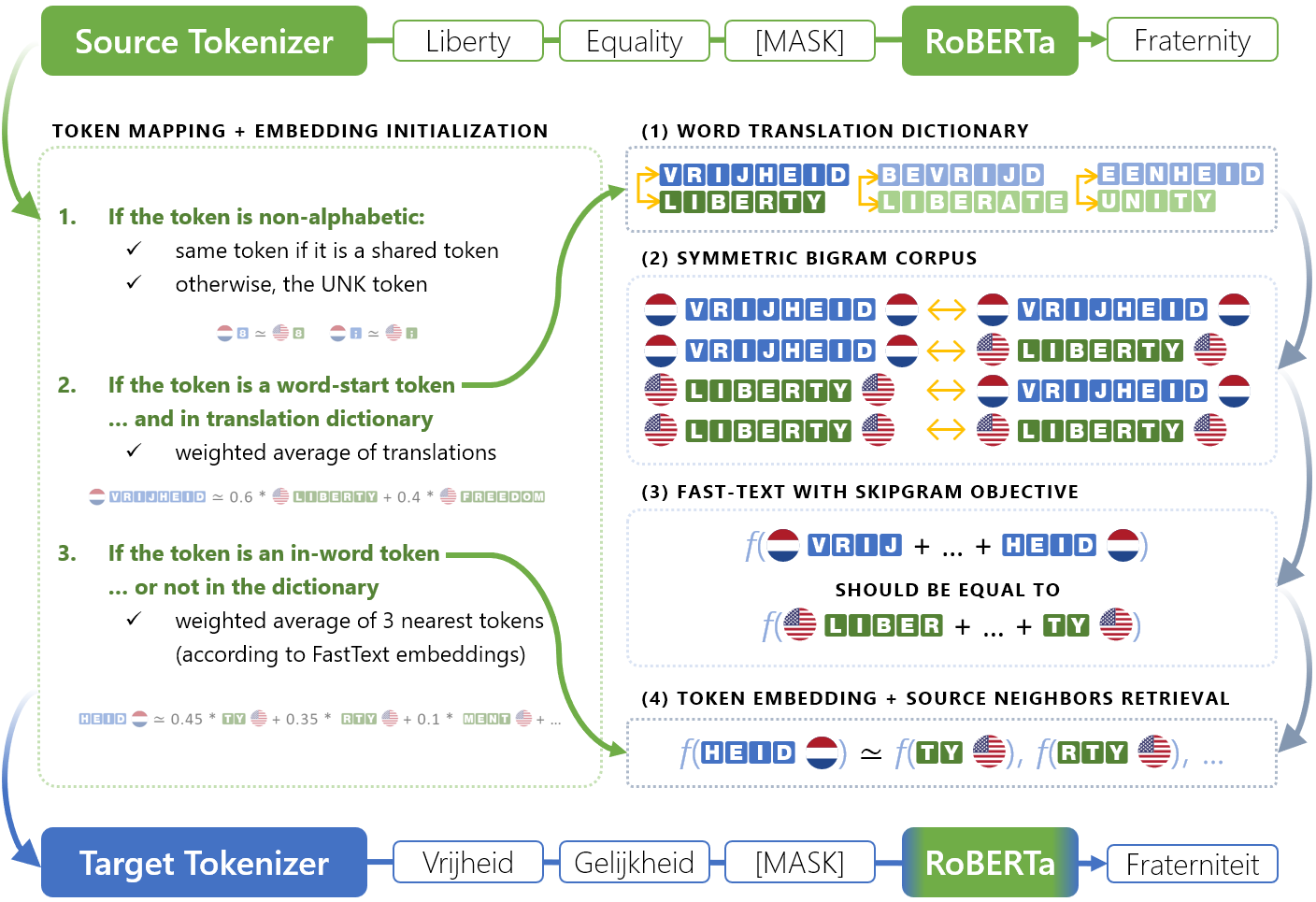}
    \caption{Illustration of the model conversion process. While non-alphabetical tokens usually have an exact match in the source tokenizer, we map the other tokens either using the word translation dictionary (1), if possible, or using a character n-gram embedding model that is able to generalize to partial tokens (4). To ensure a proper learning of the character n-gram embedding model, we generate then train on a symmetric bigram corpus (2), where every word is equally likely to be paired with itself or its translation, ensure that the skipgram distribution of a word and its translation are identical, imposing a strong equality constraint on their summed n-gram embeddings (3).\vspace*{-0.1cm}}
    \label{fig:method}
\end{figure*}

\newpage
\section{Background and Related Work }

Adapting existing language models to new languages --also called model conversion-- would be highly desirable to reduce the cost of language models. Indeed, while large language models perform exceptionally well on many downstream tasks, this performance is dependent on the high computational cost and data requirements of model pre-training, which not every language can afford.

Unfortunately, the usage of different vocabularies and tokens prevent the direct transfer of embedding weights during model conversion, requiring a random reinitialization of all or most embeddings, resulting in models of lower quality than those trained from scratch on those languages, or benefiting from multilingual transfer learning. 

Additionally, encoding input sentences using a subword tokenizer has become the dominant paradigm embraced by the state-of-the-art language models adapted from the Transformer architecture \citep{vaswani2017attention}. 
%
These tokenizers rely on a greedy approach to assign tokens to the most frequently occurring subwords by iteratively merging frequent substring pairs to form new tokens \citep{sennrich-etal-2016-neural, wu2016google}. 

Encoding words as substring tokens is undoubtedly a trade-off, whose challenges are by now well-known \citep{provilkov-etal-2020-bpe,rogers-etal-2020-primer}. One such challenge pertains to the non-morphological segmentation of compound words, as highlighted by \citet{vilar-federico-2021-statistical}.

Before explaining how we address these shortcomings, let us consider the existing strategies for low-resources languages as well as relevant work pertaining to tokenizer and model conversion.


As previously mentioned, the usage of multilingual language models remains the most common strategy used to deal with low-resource languages. Low-resources languages in multilingual models benefit from both the joint training with languages with higher-quality resources, and from a knowledge transfer obtained through to the use of a shared vocabulary \citep{pires-etal-2019-multilingual}.

One drawback of this approach is the large vocabulary it requires to properly encode all these languages, leading to an embedding table which is significantly larger than for monolingual models. 

Additionally, due to language interference, monolingual performance of mid-resourced languages only increases up to a point as more languages are added to the training, after which it starts to degrade \citep{conneau-etal-2020-unsupervised}.
The usage of multilingual language models remains therefore a trade-off for many languages worldwide, which could theoretically benefit from smaller and more powerful models, but might not do in practice due to lack of training strategies that are data-efficient enough to warrant the cost.

The perceived significance of this phenomenon seems to be increasing, as \citet{artetxe-etal-2023-revisiting} have recently demonstrated that meticulously fine-tuned multilingual model could not attain the same performance as similarly-sized monolingual English models used on translated inputs. A possible hypothesis is that it might not be possible to eliminate entirely the language interference in multilingual models, even after a monolingual finetuning.

Instead of translating inputs into English, a more sensible approach would be to convert English models into models for the target language. The first forays into the realms of model conversion originated from \citet{artetxe-etal-2020-cross}, who illustrated that performing cross-lingual transfers of monolingual models was possible, although at a significant performance cost.
Concretely, after training a new tokenizer for a target language, the authors trained a new embedding table from scratch for the new target tokens, after freezing the attention layers.

A similar approach was successfully applied on Dutch to adapt BERT~\citep{de-vries-etal-2021-adapting} and GPT-2~\citep{de-vries-nissim-2021-good}, with varying degrees of success. Their and other works demonstrated that reusing the embeddings of shared tokens could already reduce the performance gap between converted and from-scratch models.


Further investigations by \citet{gogoulou-etal-2022-cross} demonstrated that cross-lingual model conversion can, in some cases, be beneficial to model performance even in the original language, after back-conversion of the target model. This seemed to indicate it should in theory be possible to produce state-of-the-art models through conversion, if a suitable model conversion strategy was devised.




\section{Methodology}
\label{ss:methodology}

In this work, we introduce a new state-of-the-art strategy (Tik-To-Tok) to convert an existing monolingual language model from the language it was originally pretrained on (called the source language) to a new language of interest (called target). We achieve this by replacing its current tokenizer and token embedding table by new ones, which are better suited for the target language or domain. 

In line with previous works, we achieve this by reusing entries of the old embedding table to initialize the new embedding table. However, unlike previous works, we rely on a word translation dictionary spanning the source and target languages to compute a token mapping between the source and target embedding tables.

As was noted in the previous section, a better initialization of the new embedding table helps to better preserve the downstream performance of the model post conversion. However, not every shared token makes sense to reuse accross languages. While some words, such as ``\textit{computer}'' share the same meaning in Dutch and English, some other shared words such as ``\textit{of}'' have different meanings in the two languages. 

Using a word translation dictionary to perform the initialization is therefore expected to bring benefits, first by enabling many more tokens to be initialized (since being spelled identically is no longer a requirement for a better initialization), but also by reducing the frequency of inaccurate initalization where words are incorrectly initialized with the embedding of false friends.

Additionally, to address the issue of subword tokens (which are not contained as-is in a word-based translation dictionary), we devise a fallback strategy relying on an estimation of the semantic similarity of tokens. We hypothesize that it is possible to approximately translate even partial subword tokens, using an estimation of their semantic meaning based on the character n-grams they contain.

For this, we rely on fastText \cite{mikolov-et-al-2018-fasttext}, a word embedding strategy capable of producing embeddings for out-of-vocabulary subwords by summing the embeddings of the character n-gram they contain. The main advantage of fastText is its robustness against typos, according to the authors. Using fastText, we aim to provide an approximate token mapping (e.g. from \textit{Tik} to \textit{Tok}) based on the insights of the translation dictionary, even for out-of-vocabulary or for subword tokens.

To this end, we develop a bilingual fastText embedding model that generalizes over the word translation dictionary, in order to provide a mapping for all tokens not covered by the dictionary itself. We do this by embedding all source and target tokens in a shared fastText space and, for each token of the target langauge, by retrieving the tokens of the source which are its closest neighbors. 

Irrespective of the chosen strategy, we then initialize the new embedding table for each token as a weighted average of its ``translations''. We put additional emphasis on the first and second best matches, which we found to be of higher quality. To do so, we assign 30\% of the weight to the best match, 10\% to the second best match, and divide the remaining 60\% equally among all candidates.

We aim to show that this novel token translation strategy extends by a lot the potential of the translation dictionary, since many tokens (especially in Germanic languages) consist of only a part of a word or compound. We do this by comparing our strategy with state-of-the-art baselines.


\subsection{Symmetrization of the dictionary}
\label{ss:symmetrization}
To compute fastText embeddings with the desired properties, we leverage a well-known property of skip-gram models: words sharing similar neighbor distributions tend to exhibit analogous representations. Using this insight, we transform the word translation dictionary into a bigram corpus featuring a symmetric distribution (i.e., where each word is paired equally often to itself and to its translation in the other language; see Figure \ref{fig:method}.2). 

That way, a word and its translation feature exactly the same distribution of neighbors (50\% themselves, and 50\% their translation; at least for words that are perfect translations of each other) and the sum of embeddings of their character-n-grams are encouraged to match as a result (see Figure \ref{fig:method}.3). 


To differentiate words from the source and target languages, a language-specific tag is prepended and appended to every word in the dictionary (we represent these with flags in Figure \ref{fig:method}, but use simpler unicode characters in our implementation). 

For languages which form multiword compounds, like Frisian and Dutch, we perform an additional data augmentation by adding copies of each word where either the start or end tags are omitted, to simulate partial compounds. We rely on fastText to deal with extra or missing letters. 

\subsection{FastText training}
Once the symmetric bigram corpus described above is generated, we train a fastText character n-gram model on the word pairs it contains. This fastText model will be used to compute approximate semantic representations of tokens of the target language which are not present in the dictionary itself, and identify semantically similar tokens in the source language. In this section, we describe the hyperparameters used for training of the fastText model.

The default fastText training makes use of the skip-gram training objective, and we keep this default configuration in our approach. However, given that all training examples consist of pairs of words only, the skip-gram objective becomes equivalent to a neighbor prediction objective. As a result, the window size parameter no longer plays a role, and does not have to be optimized. 

To construct the embeddings of the words in the translation dictionary, fastText enumerates all the character n-grams they contain whose length is contained between \texttt{MinN} and \texttt{MaxN} characters. Each of these n-grams is then assigned an embedding, and the sum of these embeddings is used to represent the word. The default \texttt{MinN} and \texttt{MaxN} values of fastText embed all n-grams that are 3-6 characters long, but a shifted range of 4-7 characters was used in our experiments, based on the findings that 3-grams are counter-productive both for the Dutch and English languages, as was also evidenced by \citet{novotny-etal-2021-one}. 

In our experiments, we train 64-dimensional fastText embeddings on the synthetic corpus, for 5 epochs. We do not find these specific parameters to have a large impact on the resulting mapping, although training for 10 epochs or more seemed to result in an increased overfitting. 

All words from both languages are included in the fastText output space, as required by the neighbor prediction objective. All the remaining hyper-parameters are set to their default value.

\subsection{Token matching}
Once the fastText model is trained, we have all the tools necessary to perform our token mapping. \vspace*{0.15cm}

\textbf{For non-alphabetic target tokens:}
we reuse the embeddings of tokens from the source language for tokens that are shared between the two languages, without change; this mainly concerns punctuation and numeric tokens. For tokens that are not shared, the special UNK token is used as a fallback.

\textbf{For tokens present in the dictionary:} we rely on the set of translations found in the dictionary. After sorting the translations of the target word by frequency, we return the weighted average of the embeddings of all words from the list. For words in that list that do not have a corresponding token in the source tokenizer, we fallback to first token of their tokenization by the source tokenizer).

\textbf{For the remaining tokens:} we calculate their fastText embedding (after eventually prepending and/or appending the language tags, as appropriate for that token) and retrieve the three best maching source tokens using cosine similarity as a metric, and average the embedding of the three tokens. 

The strategy described above will frequently map target tokens to three or more source tokens. 
To account for the higher quality of top-ranking matches, we pre-allocate 30\% of the weight to the highest-ranking match and 10\% to the second highest-ranking match, with the residual 60\% \mbox{being} evenly divided among all candidates (including the first two). Alternatives for the weighting scheme can be investigated in future works.

Because Tik-to-Tok initializes all embeddings using the source model, our initialization strategy does not call for a random initialization unlike the previous state-of-the-art strategies.

\subsection{Embedding finetuning}
While the extensive work above yields an excellent initialization for the new embedding table of the transformer, difference in linguistic patterns between the source and target language result in a higher masked language modeling (MLM) loss in the target language after reinitialization. 

Finetuning the newly-initialized embeddings on a corpus in the target language can be used to reduce the loss again, by learning new patterns. 
To prevent catastrophic forgetting during that phase of the training, all the other parameters of the Transformer model are kept frozen.

\subsection{Model finetuning}
Once the the embeddings have converged and the MLM loss stabilizes, the base model can optionally be unfreezed in order to continue the training with the remaining data. During this second phase of the training, a lower learning rate is recommended, and a longer warmup. We provide hyperparameters for the finetuning stages in Appendix \ref{app:finetunings}.
The benefits brought by further finetuning of the whole model are analyzed in more details in \autoref{ss:dutch-experiments}.


\section{Experimental Evaluation}
\label{ss:results}

To evaluate our model conversion strategy, we conduct experiments in the low-resource and mid-resource realms of language model conversion. 
The goal of these experiments is to establish the supremacy of our proposed Tik-to-Tok initialization strategy, and benchmark its post-finetuning performance against state-of-the-art monolingual as well as state-of-the-art multilingual models.

\begin{table}[b]
\centering
\resizebox{\linewidth}{!}{%
\begin{tabular}{lccc}
\toprule
\textbf{Embedding Initialization Strategy} & \textbf{0ep.} & \textbf{1ep.} & \textbf{2ep.} \\ 
\midrule
\textbf{[1]} Random initialization \textcolor{lightgray!50!white}{\footnotesize{(Baseline)}} & 9.11 & 7.11 & 7.06 \\
\textbf{[2]} Mapping shared tokens \textcolor{lightgray!50!white}{\footnotesize{(2023 SotA)}} & 7.34 & 5.22 & \underline{5.02} \\
\textbf{[3]} Mapping with dictionary \textcolor{lightgray!50!white}{\footnotesize{(Ablation)}} & 6.56 & 4.45 & 4.23 \\
\textbf{[4]} Mapping with dictionary+fastText & \underline{5.50} & 4.00 & \textbf{3.79} \\
\bottomrule
\end{tabular}
}
\caption{MLM loss of our Dutch model after conversion into a Frisian model in function of the chosen token embedding initialization strategy (after 0, 1, and 2 epochs).}\label{table:results-frisian}
\end{table}

\begin{table*}
\caption{Results for the benchmark by \citet{delobelle2020robbert} on Natural Language Inference, Sentiment Analysis, Named Entity Recogntion, and Part-of-Speech tagging, as well as the pseudo-perplexity~\citep[PPL,][]{salazar2019mlmscoring} on the new OSCAR-2023-01 corpus. Converted models use the same tokenizer and corpus as RobBERT.
\vspace*{0.125cm}}\label{table:results-dutch}
\resizebox{\textwidth}{!}{%
\begin{tabular}{@{}rllcccccc@{}}
\toprule
\multicolumn{3}{c}{\textsc {\textbf{Configuration}}} & \multicolumn{5}{c}{\textsc {\textbf{Benchmark scores}}}                                                                                        \\ \cmidrule(lr){1-3} \cmidrule(lr){4-8}
\multicolumn{1}{c}{\textbf{Lang.}} &\textbf{Model}&  \multicolumn{1}{c}{\textbf{Params\hspace*{0.30cm}}} & \multicolumn{1}{c}{\textbf{\hspace*{0.35cm}NLI\hspace*{0.35cm}}} & \multicolumn{1}{c}{\textbf{\hspace*{0.0cm}SA\hspace*{0.0cm}}} & \multicolumn{1}{c}{\textbf{\hspace*{0.35cm}NER\hspace*{0.35cm}}} & \multicolumn{1}{c}{\textbf{\hspace*{0.0cm}POS\hspace*{0.0cm}}} & \multicolumn{1}{c}{\textbf{\hspace*{0.35cm}PPL\hspace*{0.35cm}}}\\ \midrule

\emoji{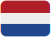} & \textcolor{darkgray}{BERTje}~\citep{devries2019bertje}
&  109~M
& 83.9 
& 93.0
& 88.3
& 96.3 
& 33.8 \\ 

\emoji{flag-netherlands} & \textcolor{darkgray}{RobBERT}~\citep{delobelle2020robbert}\hspace*{0.4cm}
& 116 M 
& 84.2 
& 94.4 
& \underline{89.1}
& \underline{96.4} 
& 13.1                         \\ 

\midrule

\hspace*{0.4cm}\emoji{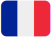}\hspace{0.1em}\pointer\hspace{0.1em}\emoji{flag-netherlands} & \textbf{Converted camembert-base}                   &          & && & & \\
& ~~~~\textcolor{gray!90!white}{Tik-to-Tok + full finetuning}
& 116 M
& 85.3 
& \underline{95.8}
& 84.9
& 94.4
& 12.4 \\

\emoji{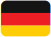}\hspace{0.1em}\pointer\hspace{0.1em}\emoji{flag-netherlands} & \textbf{Converted gbert-base}                   &          & && & & \\
& ~~~~\textcolor{gray!90!white}{Tik-to-Tok + full finetuning}
& 116 M 
& 85.5 
& 95.0
& 86.3
& 95.3 
& 10.2 \\

\emoji{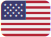}\hspace{0.1em}\pointer\hspace{0.1em}\emoji{flag-netherlands} & \textbf{Converted olm-base}
&          & && & &  \\

& ~~~~\textcolor{gray!90!white}{Tik-to-Tok only} \textcolor{lightgray}{\footnotesize(no LM head)}
& \textcolor{lightgray}{116 M}
& \textcolor{lightgray}{85.0} 
& \textcolor{lightgray}{95.5}
& \textcolor{lightgray}{78.6}
& \textcolor{lightgray}{93.8}
& \textcolor{lightgray}{{\footnotesize{$\infty$}}} \\

& ~~~~\textcolor{gray!90!white}{Tik-to-Tok + embeddings ft.}
& \textcolor{lightgray}{116 M}
& \textcolor{lightgray}{85.4}
& \textcolor{lightgray}{95.6}
& \textcolor{lightgray}{86.0}
& \textcolor{lightgray}{95.1}
& \textcolor{lightgray}{9.9} \\

& ~~~~\textcolor{gray!90!white}{Tik-to-Tok + full finetuning}
& 116 M
& \underline{86.6} 
& 95.4
& 87.6
& 95.8 
& \underline{5.9} \\

\emoji{flag-united-states}\hspace{0.1em}\pointer\hspace{0.1em}\emoji{flag-netherlands} & \textbf{Converted roberta-large} 
&          & && & & \\
& ~~~~\textcolor{gray!90!white}{Tik-to-Tok + full finetuning}                  & 345 M           & $ \textbf{89.2} $ & \textbf{97.0} $ $                   & $\textbf{89.5}$             & $96.0 $ & \textbf{4.9}      \\

\midrule

\emoji{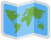} & \textcolor{darkgray}{XLM-RoBERTa large (XLM-R)}        & 560 M           & 87.9 & {96.5}  & \textbf{89.5} & \textbf{96.9}                           & 5.5                   \\ 

  \bottomrule

\end{tabular}%
}
\end{table*}

\subsection{Low-resource Languages}\label{ss:contrib-1}

To compare our embedding table initialization strategy with state-of-the-art approaches, we first focus on experiments on a low-resource language: Frisian, a West Germanic language spoken by about 400,000 native speakers, mostly located in the province of Friesland (Fryslân) in the north of the Netherlands. The grammar of the Frisian language is similar to other West Germanic languages and most notably the Dutch language.

The Oscar 2019 corpus contains only 35 megabytes of data for Frisian, making it impossible to training large language models for Frisian. However, its proximity with Dutch enabled \citet{de-vries-etal-2021-adapting} to convert an existing Dutch model into a mostly functional Frisian model.

In this work, we take advantage of the very small size of the Frisian corpus used in that experiment to quantify the impact of different embedding initialization strategies on the conversion outcome. 

In particular, we investigate four initialization strategies: 
\textbf{[1]} a complete re-initialization of the embedding table (baseline), 
\textbf{[2]} a re-utilization of the source embeddings of all tokens which are shared between the Dutch and Frisian tokenizers (previous state of the art), 
\textbf{[3]} a simplified dictionary mapping strategy where only non-alphabetic tokens and full-word tokens present in the word translation dictionary are initialized based on the source embeddings, and finally
\textbf{[4]} our complete Tik-To-Tok mapping strategy where gaps in the dictionary mapping are filled using the fastText embeddings computed for source and target tokens.

In this experiment, we convert our \textit{roberta-large-nl-oscar19} model (described in the next section) into a Frisian model using these various strategies. We therefore apply our mappings between the Dutch tokenizer of RobBERT \citep{delobelle2020robbert}, and the Frisian tokenizer devised by \citet{de-vries-etal-2021-adapting}.

The model parameters are kept frozen throughout our experimentation, to the exception of the language modeling head and the embedding table. The limited scope of the parameter training is justified by the data scarcity, the high grammatical proximity between the Frisian and Dutch languages, and the strong cultural and geographical ties between the two communities. 

To create a word translation dictionary suitable for the token mapping strategies that require it, and given that the Frisian language is not supported by the OpenSubtitles2018 dataset \citep{lison-etal-2018-opensubtitles2018}, we combine two manually curated word translation dictionaries between Frisian and Dutch, from respectively \citet{duijff-etal-2008-frysk} and \citet{zantema-1984-frysk}.

We evaluate the models based on the MLM loss after 0, 1 and 2 training epochs on the oscar corpus. This evaluation strategy was chosen to show both differences in post-initialization quality (0 epoch) and the extent to which this difference can be recovered through finetuning using the available Frisian corpus.
The loss after 2 training epochs is reported to demonstrate that a plateau is already reached after 1 training epoch, and that little further performance gain is to be expected beyond that point (without providing significantly more training data).
Our results are reported in Table \ref{table:results-frisian}.

From these results, we draw the following conclusions: 
\textbf{[a]} Irrespective of the conversion strategy, finetuning on the native corpus showed limited \textcolor{gray}{\footnotesize(\textasciitilde2p)} capacity for recovering from the impairment caused by the conversion, highlighting the need of a good initialization. 
\textbf{[b]} Reusing shared tokens is a massive improvement over the random re-initialization of the embedding table, but the final performance remains subpar. 
\textbf{[c]} Reusing embeddings through a dictionary mapping, on the other hand, results in a much better model at initialization; the MLM loss of our Tik-to-Tok model is already comparable with the loss of a finetuned model converted using the SOTA approach (with MLM finetuning bringing additional improvements on top of that).

\subsection{Mid-resource Languages}\label{ss:contrib-2}
\label{ss:dutch-experiments}
Even whenever enough data is available to pre-train a language model, our token translation approach can significantly reduce the time and cost required for pre-training such a language model.
We demonstrate this by applying our Tik-to-Tok model conversion to Dutch, a mid-resource language.

We train a series of Dutch models, on the same corpus (Oscar19 NL) and using the same tokenizer as RobBERT \citep{delobelle2020robbert}, then evaluate them following the same evaluation as this existing Dutch model. We also compare them with BERTje \citep{devries2019bertje} and the (larger) multilingual XLM-R model \citep{conneau-etal-2020-unsupervised}.

We evaluate a set of high-resource languages (French, German, and English) to initialize our Dutch models. More specifically we evaluate converted versions of the French CamemBERT-base \citep{martin-etal-2020-camembert}, the German GBERT-base \citep{chan-etal-2020-germans}, and the English olm-base \citep{thrush-and-oblokulov-2022-olm} and RoBERTa-large \citep{liu2019roberta} models. Three different tokenizers (BertTokenizer, RobertaTokenizer, CamembertTokenizer) are covered by these four models, thereby proving that our approach can be used across a wide range of tokenizer implementations.

The models are evaluated on 5 tasks: Sentiment Analysis (SA), Named Entity Recognition (NER), Part-of-Speech tagging (POS), Natural Language Inference (NLI), as well as their Pseudo-Perplexity (PPL) on the recently-crawled Oscar2023 corpus.
\newpage
\vspace*{0.15cm}
To disentangle the effects of the embedding table initialization from the subsequent MLM finetuning, we perform an ablation study where we test three variations of the training setup.
Due to the limited impact, this seemed to have on the downstream tasks, we only performed this analysis for our best 116M-parameters model, initialized from olm-base.
Results for the other models are presented only after finetuning the weights of the entire model.

In all cases, we train the models following the same procedure:
Firstly, we reinitialize the embeddings of the transformer model following the soft token-mapping procedure described in \autoref{ss:methodology}, using the RobBERT tokenizer as target. Secondly, we finetune the newly-initialized embeddings and the language modeling head on a corpus of the target language. Finally, we unfreeze all the weights and continue finetuning on the same corpus.

We stop all experiments early, after using only about 15\% of the available Dutch data (\textasciitilde7Gb text) of the Oscar19 corpus, because the training loss stabilizes around that point; another experiment using 25\% of the available Dutch data, not reported in the table, showed no further improvement.

Our results, summarized in \autoref{table:results-dutch}, seem to indicate that models initialized with high-resources languages perform better and train faster than state-of-the-art language models, with our best model outperforming all alternatives in 4 out of the 5 tasks, while being competitive on the remaining one. 

Our ablation study confirms that while finetuning the embeddings and the transformer weights on a Dutch corpus improves the performance on the downstream tasks measurably, this is not required for achieving state-of-the-art performance in most of the tasks considered in our benchmark, expanding on our observation that our initialization performs well as-is for low-resource languages. This is very promising, as it is difficult to finetune the transformer weights for many low-resource languages due to lack of available corpora (see \autoref{ss:contrib-1}).

Unlike what is reported in \citet{de-vries-etal-2021-adapting}, we find little evidence that high language similarity is critical for downstream task performance, with our Romance-initialized model (camembert-base) and our German-initialized model (gbert-base) performing about equally well on average. One notable exception to this observation concerns part-of-speech tagging, a fine-grained grammatical task; interestingly, this is the task on which \citeauthor{de-vries-etal-2021-adapting} evaluated their converted Frisian~model, which probably explains their conclusion.

%
%
%
%
%





\section{Impact and Discussion}
We believe the results presented above could have a large impact on the NLP community, by making the creation of high-quality models for low-resource languages more accessible. While this work focused on general-purpose models, the conversion of more specialized models, such as 
biomedical models, could be envisioned 
using specialized translation dictionaries \cite{remy-etal-2022-translating}.

Another interesting aspect of our approach is its potential for model distillation \citep{hinton2015distilling,bucilua2006modelcompression}. In our second experiment, several high-resource language models were converted to Dutch models, using the same tokenizer. The usage of different tokenizers is one of the key reasons why language models are usually difficult to combine with each other, an issue that our technique can contribute to overcome. 

Finally, because converting models using our technique is inexpensive, researchers using our methodology might be able to update language models on a more regular basis, an idea which the Online\,Language\,Modeling\,community contributed to popularize \cite{thrush-and-oblokulov-2022-olm}.

Our code and models will be released upon acceptance. 
We provide an easy-to-use notebook to convert and finetune language models for any of the 1782 language pairs supported by OpenSubtitles. 

\section{Conclusion}
In this work, we were able to successfully improve the performance of language model conversion by introducing a new token translation task. \vspace{0.1cm}

We demonstrated how this new initialization strategy largely benefits low-resource languages such as Frisian, but also makes it possible to train monolingual models for languages with more resources (such as Dutch) using far less training data and time as was previously possible.  \vspace{0.1cm}

Our results show that this approach might improve the state-of-the-art tools available for many languages across the globe, a key fairness issue.\vspace{0.1cm}

We also believe our work is opening the gate to more frequent incremental updates of these models to keep up with the changing patterns of language over time, an often overlooked problem.\vspace{0.2cm}

\section{Limitations and Future Work}

While this project features exciting development for low- and mid-resource languages, a couple of limitations of our work are worth discussing.

One of these limitations is that our experiments only cover Romance and Germanic languages. Collaborations to work on a more diverse set of languages are being discussed, but are at early stages.

Another limitation is that converting models to other languages might have unexpected effects on the model's linguistic and cultural understanding. This might for instance exacerbate or dampen the biases present in the training data. 

The impact of the use of a particular word translation dictionary, or the combination from multiple dictionaries, was not studied in this work, but might be important to consider as well. Using word translation dictionaries derived from aligned subtitles, like we do for Dutch, might possibly bias the language model understanding towards certain topics.

Finally, because our work relies on a token-level translation task, the understanding of multiword expressions in the target language is also a subject of concern. This understanding will probably have to be learned through the full finetuning of the model, as token embeddings are unlikely to be sufficient to learn them all in isolation. Even when the translation dictionary contains some multiword expressions, we do not provide a way to use them effectively in our proposed framework. Future works might want to provide an extension of 
our strategy to multiword expressions.



\FloatBarrier
\newpage
\bibliography{anthology,custom,custom_local}
\bibliographystyle{acl_natbib}


\clearpage
\appendix
\section{Model conversion details}
\label{app:finetunings}
In this appendix, we provide details necessary to replicate our Dutch model finetuning, as the hyperparameters required for this task are numerous, and not always relevant to mention in the main text.

Our finetuning was divided into three phases: \textbf{[1]} an embedding finetuning, focused on improving the initialization outcome, \textbf{[2]} a grammatical finetuning, focused on learning and unlearning languistic patterns in the edge layers of the Transformer, and \textbf{[3]} a knowledge finetuning, where all the weights of the Transformer are finetuned.

As the number of tunable parameters varies, so must the learning rate and warmup period, to ensure high-quality results.

\subsection{Embedding finetuning}
During this phase, around 5\% of the Dutch corpus was used for training. In this phase, only the embedding weights and the language modeling head are tunable parameters.
\begin{verbatim}
    per_device_train_batch_size=4
    gradient_accumulation_steps=8, 
    #total_batch_size=32,
    training_steps=150000,
    learning_rate=5e-5,
    warmup_steps=5000,
    weight_decay=0.01,
    fp16=True,
\end{verbatim}

\subsection{Grammatical finetuning}
In the second step, around 5\% of the Dutch corpus was used for training. The main change over the previous phase are the unfreezing of the bottom two and top two Transformer layers, and the increase of the batch size for less noisy gradients.
\begin{verbatim}
    per_device_train_batch_size=4
    gradient_accumulation_steps=64, 
    #total_batch_size=256,
    training_steps=25000,
    learning_rate=5e-5,
    warmup_steps=1000,
    weight_decay=0.01,
    fp16=True,
\end{verbatim}

\newpage
\subsection{Knowledge finetuning}
In the second step, around 5\% of the Dutch corpus was used for training. The main change over the previous phase are the unfreezing of all remaining Transformer layers, and the decrease of the learning rate to adjust to the increase of non-linear effects in the parameter updates.
\begin{verbatim}
    per_device_train_batch_size=4
    gradient_accumulation_steps=64, 
    #total_batch_size=256,
    training_steps=25000,
    learning_rate=2e-5,
    warmup_steps=2000,
    weight_decay=0.01,
    fp16=True,
\end{verbatim}

\subsection{Summary}
In total, about 15\% of the training data were used during the full finetuning procedure. This amounts to about 7Gb of text (out of the 47Gb available in the Oscar corpus).

We performed the tuning on a single V100 GPU, with a total running time of about a week for each model trained.

\clearpage

\section{Dutch evaluation details}
\label{app:setup}

In this section, we provide more details about the Dutch evalution performed in section \ref{ss:contrib-2}.
\subsection{Sentiment Analysis (SA)}
We evaluate sentiment analysis on the Dutch Book Review Dataset~\citep{vanderburgh2019dbrd} with standard splits.
This dataset is publicly available with a cc-by-nc-sa-4.0 licence.
Our experiment consists of one run with the following hyperparameters:

\begin{itemize}
  \setlength\itemsep{0.3ex}
    \item Number of gpus: 1 (1080 Ti)
    \item adafactor: False
    \item adam beta1: 0.9
    \item adam beta2: 0.999
    \item adam epsilon: 1e-08
    \item deepspeed: None
    \item fp16: False
    \item gradient acc. steps: 8
    \item lr: $ 10^{-4}$
    \item lr scheduler type: LINEAR
    \item num train epochs: 10
    \item optimizer: ADAMW
    \item batch size: 4
    \item seed: 1
    
    \item warmup ratio: 0.0
    \item warmup steps: 20
    \item weight decay: 0.05
    
\end{itemize}

\subsection{Named Entity Recognition (NER)}
We evaluate NER on the CoNLL-2002 shared task from \url{https://www.clips.uantwerpen.be/conll2002/ner/} (no explicit mention of a licence) with an experiment that consists of 10 runs with Bayesian optimisation (TPE) with the following hyperparameters, where we vary the learning rate, number of gradient accumulation steps and weight decay. We select the best-performing model based on the $F_1$ score on a separate validation set before testing this model the test set.

\begin{itemize}
  \setlength\itemsep{0.3ex}
    \item Number of gpus: 1 (1080 Ti)
    \item adafactor: False
    \item adam beta1: 0.9
    \item adam beta2: 0.999
    \item adam epsilon: 1e-08
    \item deepspeed: None
    \item fp16: False
    \item gradient acc. steps: $\{1, 2, 4, 8, 16, 32\}$. 
    \item lr: $[10^{-6},10^{-4}]$. 
    \item lr scheduler type: LINEAR
    \item num train epochs: 10
    \item optimizer: ADAMW
    \item batch size: 8    
    \item warmup ratio: 0.0
    \item warmup steps: 20
    \item weight decay: $[0.01, 0.1]$.
    
\end{itemize}

\subsection{Part-of-speech (POS) tagging}
We used the Dutch part of the Lassy corpus~\citep{bouma-van-noord-2017-increasing} available at \url{https://universaldependencies.org/treebanks/nl_lassysmall/index.html} which has a cc-by-sa 4.0 licence. 
We perform 10 runs with Bayesian optimisation (TPE) with the following hyperparameters, where we vary the learning rate, number of gradient accumulation steps and weight decay. We select the best-performing model based on the $F_1$ score on a separate validation set before testing this model the test set. 

\begin{itemize}
  \setlength\itemsep{0.3ex}
    \item Number of gpus: 1 (1080 Ti)
    \item adafactor: False
    \item adam beta1: 0.9
    \item adam beta2: 0.999
    \item adam epsilon: 1e-08
    \item deepspeed: None
    \item fp16: False
    \item gradient acc. steps: $\{1, 2, 4, 8, 16, 32\}$. 
    \item lr: $[10^{-6},10^{-4}]$.
    \item lr scheduler type: LINEAR
    \item num train epochs: 10
    \item optimizer: ADAMW
    \item batch size: 8
    \item warmup ratio: 0.0
    \item warmup steps: 20
    \item weight decay: $[0.01, 0.1]$.
\end{itemize}

\subsection{Natural Language Inference (NLI)}
We use the SICK-NL dataset \citep{wijnholds2021sicknl}, available at \url{https://github.com/gijswijnholds/sick_nl} under the MIT licence.
Our experiment consists of 30 runs with the following hyperparameters, where most are fixed and the learning rate, weight decay and the number of gradient accumulation steps are randomly selected from the specified ranges.

\begin{itemize}
  \setlength\itemsep{0.3ex}
    \item Number of gpus: 1 (1080 Ti)
    \item adafactor: False
    \item adam beta1: 0.9
    \item adam beta2: 0.999
    \item adam epsilon: 1e-08
    \item deepspeed: None
    \item fp16: False
    \item gradient acc. steps: $\{2, 4, 8, 16\}$.
    \item lr: $[10^{-6},10^{-4}]$.
    \item lr scheduler type: LINEAR
    \item num train epochs: 10
    \item optimizer: ADAMW
    \item batch size: 8
    \item warmup ratio: 0.0
    \item warmup steps: 20
    \item weight decay: $[0, 0.1]$.
    
\end{itemize}

\end{document}